%% file: conference.tex
\documentclass[conference]{IEEEtran}
\IEEEoverridecommandlockouts
\usepackage{cite}
\usepackage{amsmath,amssymb,amsfonts}
\usepackage{algorithmic}
\usepackage{graphicx}
\usepackage{textcomp}
\usepackage[table]{xcolor}
\usepackage{multicol}
\usepackage{subcaption}
\usepackage{listings}
\usepackage{tikz}

\def\BibTeX{{\rm B\kern-.05em{\sc i\kern-.025em b}\kern-.08em
    T\kern-.1667em\lower.7ex\hbox{E}\kern-.125emX}}
\begin{document}

\title{Towards making the most of NLP-based device mapping optimization for OpenCL kernels\\
}

\definecolor{codegreen}{rgb}{0,0.6,0}
\definecolor{codegray}{rgb}{0.5,0.5,0.5}
\definecolor{codepurple}{rgb}{0.58,0,0.82}
\definecolor{backcolour}{rgb}{0.95,0.95,0.92}
\lstdefinestyle{mystyle}{
    backgroundcolor=\color{backcolour},   
    commentstyle=\color{codegreen},
    keywordstyle=\color{magenta},
    numberstyle=\tiny\color{codegray},
    stringstyle=\color{codepurple},
    basicstyle=\ttfamily\footnotesize,
    breakatwhitespace=false,         
    breaklines=true,                 
    captionpos=b,                    
    keepspaces=true,                 
    numbers=left,                    
    numbersep=5pt,                  
    showspaces=false,                
    showstringspaces=false,
    showtabs=false,                  
    tabsize=2
}


\author{\IEEEauthorblockN{Petros Vavaroutsos, Ioannis Oroutzoglou, Dimosthenis Masouros, Dimitrios Soudris}
\IEEEauthorblockA{\emph{
School of Electrical and Computer Engineering,
National Technical University of Athens,
Greece
} \\
Emails:
\{petrosvav,
ioroutzoglou,
demo.masouros,
dsoudris\}@microlab.ntua.gr}
}

\DeclareRobustCommand\circled[1]{\tikz[baseline=(char.base)]{
            \node[shape=circle,fill,inner sep=0.4pt] (char) {\textcolor{white}{#1}};}}

\IEEEoverridecommandlockouts
\IEEEpubid{\makebox[\columnwidth]{978-1-6654-8356-8/22/\$31.00~\copyright2022 IEEE
\hfill} \hspace{\columnsep}\makebox[\columnwidth]{ }}
\maketitle
\IEEEpubidadjcol

\begin{abstract}
Nowadays, we are living in an era of extreme device heterogeneity.
Despite the high variety of conventional CPU architectures, accelerator devices, such as GPUs and FPGAs, also appear in the foreground exploding the pool of available solutions to execute applications.
However, choosing the appropriate device per application needs is an extremely challenging task due to the abstract relationship between hardware and software.
Automatic optimization algorithms that are accurate are required to cope with the complexity and variety of current hardware and software. 
Optimal execution has always relied on time-consuming trial and error approaches. Machine learning (ML) and Natural Language Processing (NLP) has flourished over the last decade with research focusing on deep architectures. In this context, the use of natural language processing techniques to source code in order to conduct autotuning tasks is an emerging field of study. 

In this paper, we extend the work of Cummins et al., namely Deeptune, that tackles the problem of optimal device selection (CPU or GPU) for accelerated OpenCL kernels.
We identify three major limitations of Deeptune and, based on these, we propose four different DNN models that provide enhanced contextual information of source codes.
Experimental results show that our proposed methodology surpasses that of Cummins et al. work, providing up to 4\% improvement in prediction accuracy.
\end{abstract}


\input{sections/1_intro}

\input{sections/related}
\input{sections/background}
\input{sections/methodology}
\input{sections/evaluation}
\input{sections/conclusion}



\bibliographystyle{IEEEtranS}
\bibliography{conference}

\end{document}

%% file: sections/1_intro.tex
\section{Introduction}
\label{sec:intro}

The ever-increasing amount of data generated and shared by enterprises, industrial and non-profit sectors, and scientific research has resulted in an unprecedented increase in the size and volume of data-intensive jobs~\cite{Agarwal2014big}.
In order to meet the new computational demands of the big data, both hardware and software undergo major changes.
Additionally, new latency and power constraints require approaches to solve number of different application and compiler optimization problems. 
However, their large design decision space to explore, makes tuning applications an even more difficult and time consuming procedure, making the development of tuning heuristics more urgent than ever.
Furthermore, contemporary compilers and runtime environments, featuring already hand-coded heuristics, performing this decision making, make their program's performances contingent upon their quality.   


Therefore, in order to make heuristic construction more efficient and inexpensive, automation tools have to be imported in this procedure.
In this regard, machine learning techniques have been deployed in order to automate the process of selecting the best optimizations~\cite{ashouri2018survey}.  
Prediction models using different classes of machine learning are trained through applications' representative features in order to correlate them with their optimal versions.
Features can be any important quantifiable properties of applications, static or dynamic, and the choice of the most appropriate features constitutes an extra optimization problem for the designers as well. 

Even though machine learning has a proven contribution in automated tuning heuristics~\cite{grewe2013portable, magni2014automatic}, its success is contingent on the quality of the extracted features, which is frequently achieved through a combination of domain expertise and trial and error.
In order to avoid the extraction of non appropriate features and finally inefficient tuning models, humans are needed to be removed from the loop.
Latest publications~\cite{cummins2017end, ben2018neural} focus their work on characterizing and tuning applications without using any code feature.
Natural Language Processing (NLP) methods are deployed to extract automatically and internally code's text features and feed them to the predictive model.
Therefore, NLP models are now able to extract feature representations from source codes automatically and afterwards, other learning systems could employ these learnt feature representations as inputs to predict various down stream tasks.

Among the dozens of optimizations, lately, most of them are focused on the accelerators devices (such as GPUs and FPGAs) deployed to satisfy the even more demanding performance and power constraints from the edge to cloud continuum.
One of the most effective and popular one, concerns the optimal heterogeneous device mapping optimization for OpenCL written applications.
More specifically, this optimizations refers to the selection of either CPU or GPU device for the most efficient execution of OpenCL kernels in terms of performance.

In this work, we extend DeepTune~\cite{cummins2017end}, one of the most influential works on machine-learning based auto-tuning methodologies without any hand engineered feature extractors, attempting to solve the heterogeneous device mapping optimization.
We achieve to further improve its effectiveness and finally provide more efficient auto-tuning methods without any need for feature engineering.
Our proposed work outperforms DeepTune by providing up to 4.12\%, with average 2.65\% higher prediction accuracy for the optimal device mapping selection.
The rest of paper is structured as follows: Section II introduces the related work already published concerning the automation of applications tuning methods.
Section III describes the overview of DeepTune's baseline implementation while in Section IV we present our proposed improved methodology.
The experiments and the findings are described in Section V and finally, in Section VI we draw our conclusions.

%% file: sections/related.tex
\section{Related Work}
\label{sec:related}

There have been numerous works conducting with the goal of optimizing source code using machine learning models in order to automate tuning procedure \cite{allamanis2018survey}.
Back in the early 2008, Agakov et al.~\cite{agakov2006using} were the first to use a machine learning based predictive model to speed up iterative optimization while, in the same year, Cavazos et al.~\cite{cavazos2006method} applied logistic regression models to automatically select the best optimization for user's applications.
Later, Liu et al.~\cite{liu2009cross} turning to accelerators, used regression trees to optimize CUDA kernels, while Cummins et al.~\cite{cummins2015autotuning} applied classifiers to select the optimal workgroup size of OpenCL compute kernels.
Finally, Magni et al.~\cite{magni2014automatic} were the first to address the coarsening optimizations through Neural Networks cascade models.  


Even though the above works managed to provide sufficient results based on hand crafted program features by developers, heuristics needed to take humans out of the loop.
Building auto-tuning models without feature engineering can provide faster, cheaper and more independent heuristics achieving to discover more optimal tuning solutions without any human guidance. 
Cummins et al.~\cite{cummins2017end} where the first to introduce their work in that direction, with deep neural networks to replace any hand-picked or even compiler IR based automatically extracted features.
Surprisingly, their approach matched or surpassed the predictive models using hand-crafted features, proving that deep learning can select more representative and sufficient features than even experts engineers.
In the same direction, Ben-Nun et al. ~\cite{ben2018neural} managed to apply NLP techniques on the Intermediate Representations (IR) of applications in order to take the feature extraction out of the loop to support a wider range of programming languages.
Their experimental results were encouraging, disclosing that the hand-coded features, are not, but an obstacle in the building of the auto-tuning heuristics.


%% file: sections/background.tex
\section{Deeptune Overview}
\label{sec:deeptune}

 \begin{figure*}[ht]
    \centering
    \includegraphics[width=\textwidth]{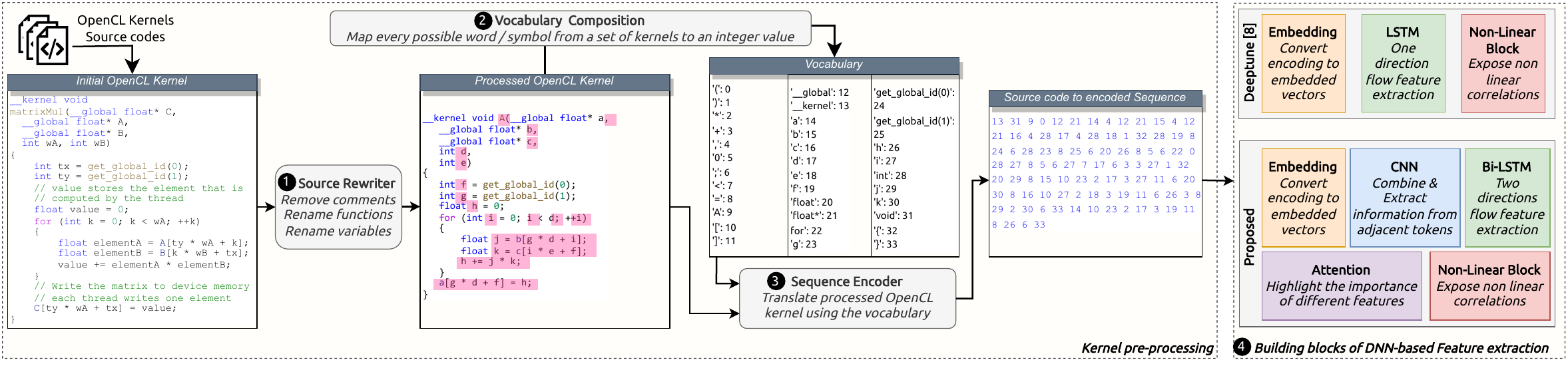}
    \caption{High-level overview of the end-to-end optimal device mapping methodology for OpenCL kernels. The \textit{kernel-preprocessing} phase transforms source codes to machine learning interpretable format and the \textit{feature extraction}, based on the optimization problem, it generates highly descriptive characteristics for each kernel. \circled{4}-top shows the \textit{feature extraction} approach of Deeptune \cite{cummins2017end} and \circled{4}-bottom our proposed building blocks.}
    \label{fig:methodology-overview}
\end{figure*}

In this paper, we base our research in the work of~\cite{cummins2017end}. 
Cummins et. al present an end to end architecture that accepts OpenCL kernels as input and decides on the optimal device that those kernels should execute, CPU or GPU. 
Their proposed methodology consists of two phases: \textit{i)} a kernel pre-processing stage, that transforms the OpenCL kernels to a machine-learning friendly interprentetation and \textit{ii)} a DNN-based feature extraction phase, that receives the transformed code and based on its features identifies the more performance efficient device mapping for the respective kernel.
Figure \ref{fig:methodology-overview} shows an overview of the kernel transformation flow, as well as the basic layers, used as building blocks for feature extraction used in~\cite{cummins2017end}.
Next, we briefly describe the basic steps of the work of Cummins et. al, henceforth referred to as Deeptune.

\subsection{Source Rewriter}
\label{subsec:source-rewriter}
The first stage of Deeptune consists of a source rewriter \circled{1}.
Its purpose is to reconstruct a given OpenCL kernel to a refined version that  eliminates semantically irrelevant information.
This refined version can then be more easily processed in an automated manner.
Specifically, this component accepts hand written code and it performs the following three actions \textit{i)} removes comments and erase unnecessary spacing \textit{ii)} rewrites function names using an increasing alphabetical order of capital letters [A-Z] and \textit{iii)} rewrites the codes variable using an increasing alphabetical order of lowercase letters [a-z].

\subsection{Vocabulary Composition}
\label{subsec:vocabulary-composition}
The second step of Deeptune's methodology is the vocabulary composition \circled{2}.
This step concerns the construction of a vocabulary that maps code related definitions (e.g., \texttt{\_\_kernel}, \texttt{int}) and punctuation marks (e.g., parentheses, semicolons) to a set of integer representations.
This vocabulary is mandatory, since machine-learning models receive as input numeric values and are not able to process source code directly.
To build the corpus vocabulary, this step considers all the possible words and symbols appeared over all the examined OpenCL kernels and performs word level tokenization, which maps each token to a unique integer identifier, called \textit{token id}.

\subsection{Sequence Encoder}
\label{subsec:sequence-encoder}
The last step of the pre-processing phase is the sequence encoder \circled{3}.
This component combines the processed OpenCL kernel (output of step \circled{1}) with the corpus vocabulary (output of step \circled{2}) and provides an integer sequence that corresponds to the respective input code.

\subsection{DNN-based feature extraction}
As mentioned before, Deeptune utilizes a DNN-based approach to automate the process of feature extraction, thus eliminating completely the need for hand-crafted solutions.
The proposed DNN architecture consists of three major layers, that form the building blocks of their model (\circled{4}--top).
First, an Embedding layer acts as a language model that receives as input the tokenized source code sequence and converts them to embedding vectors of dimension $D=64$.
Next, the embedded vector is passed to a block of two Long Short-Term Memory~\cite{hochreiter1997long} layers, that is responsible for exposing one-way sequential information on the input vector.
Last, DeepTune's model final component is a non-linear block, that is responsible for exposing non-linearities in the data.
It consists of two fully connected artificial neural network layers, where the initial layer is composed of 32 neurons. Each possible heuristic decision is represented by a single neuron in the second layer. The model's confidence that the associated choice is right is represented by the activation of each neuron in the output layer. Taking the output layer's argmax yields the decision with the highest activation.

\subsection{Deeptune's Limitations \& Motivation}
\label{subsec:deeptune-limitations}
While the work of~\cite{cummins2017end} is revolutionary in automating the feature extraction process of GPU code, we argue that it neglects important aspects and inherent characteristics of the structure and nature of modern programming languages, which, if approached properly, could provide useful insights.
Moreover, novel advancements in the domain of machine learning and NLP~\cite{vaswani2017attention,ben2018neural,devlin2018bert},
can provide more representative and/or diverse features, unveiling additional hidden patterns in the underlying data.
Specifically, we pinpoint three important remarks that Deeptune disregards and which also form the motivation of our work:

\textbf{R1) Relation between adjacent tokens:} The structure of a programming language is very well defined, with syntactical rules. The syntax of a programming language is a collection of rules defining the combinations of symbols that constitute appropriately organized statements or expressions in that language. Deeptune neglects that structure, by treating tokens sequentially, thus it makes no consideration for the premise that adjacent tokens provide additional context for code comprehension. In the example below, the model will assign a high probability to the right token, which is \texttt{int}, based on the structure of the variable's \texttt{a} declaration.
\begin{lstlisting}[style=mystyle,language=C,caption=Example code that highlights the dependence of adjacent tokens]
int main(){
    int a = 10;
    ___ b = a + 1;
}
\end{lstlisting}

\textbf{R2) Preserve past and future information:} While Deeptune examines hidden relationships in the sequential data through the LSTM layers, it only acquires knowledge exclusively from the input's subsequent pass, since unidirectional LSTMs only preserve inputs that has already passed through it using the hidden state.
However, the typical flow of imperative programming reveals bidirectional inter-dependencies, from the beginning of the code to its end and vice-versa.
A typical example is the following: 
\begin{lstlisting}[style=mystyle,language=C,caption=Example code that depicts future dependencies]
float sqrt(int a);
int main(){
    int x = sqrt(2); 
    // several lines of code
}
float sqrt(int a){
    // implementation
}
\end{lstlisting}
where the call of a function is several lines afterwards than it's implementation.

\textbf{R3) Significance-aware feature processing:} Last, Deeptune treats all the input features equally, even if part of it is less significant.
However, it is apparent that not all source code semantics provide proportionate contextual information.
For example, a variable declaration (e.g., \texttt{int}) is not as critical as the definition of a loop (e.g., \texttt{for}).
Therefore, it is of great importance to be able to distinguish more valuable features from the less significant ones.


%% file: sections/methodology.tex
\section{Proposed Methodology}
\label{sec:methodology}
This section describes the core concepts of our methodology for optimal device mapping of OpenCL kernels between CPU and GPU devices based on NLP techniques.
The rationale behind the proposed processing flow is driven by the remarks (R1-R3) drawn in Section~\ref{subsec:deeptune-limitations}.
Specifically, similar to~\cite{cummins2017end}, our methodology consists of two phases, the \textit{kernel pre-processing} and the \textit{feature extraction}, as depicted in Fig.~\ref{fig:methodology-overview}.
However, it extends the latter to account for additional contextual information on the examined kernels.

\subsection{Kernel pre-processing}
The kernel pre-processing phase of our methodology is identical to the one proposed in Deeptune.
We replicate the pre-processing components described in Sections~\ref{subsec:source-rewriter} to~\ref{subsec:sequence-encoder}, which correspond to steps \circled{1}-\circled{3} of Fig. \ref{fig:methodology-overview}.

\begin{figure*}[ht]
    \centering
    \subfloat[Deeptune \cite{cummins2017end}]{
        \centering
        \includegraphics[width=0.15\textwidth]{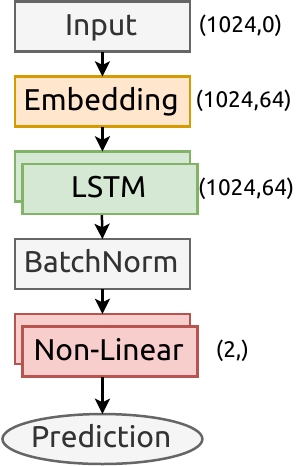}
        \label{fig:deeptune}
    }
    \hfill
    \subfloat[Deeptune-CNN]{
        \centering
        \includegraphics[width=0.15\textwidth]{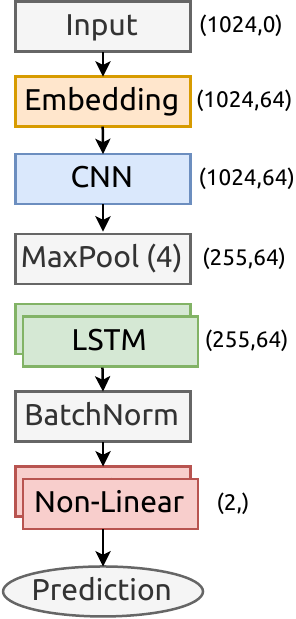}
        \label{fig:cnn}
    }
    \hfill
    \subfloat[Deeptune-BiLSTM]{
        \centering
        \includegraphics[width=0.15\textwidth]{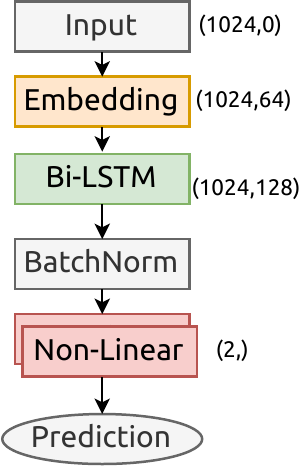}
        \label{fig:bilstm}
    }
    \hfill
    \subfloat[Deeptune-Attention]{
        \centering
        \includegraphics[width=0.15\textwidth,scale=.8]{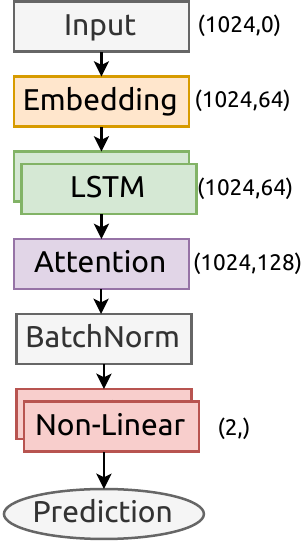}
        \label{fig:attention}
    }
    \hfill
    \subfloat[Hybrid Architecture]{
        \centering
        \includegraphics[width=0.15\textwidth,scale=.8]{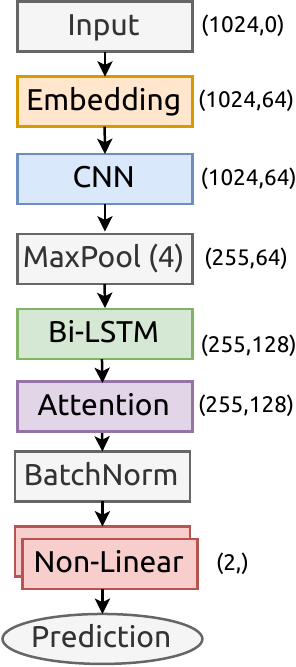}
        \label{fig:proposed}
    }
    \caption{Examined DNN architectures}
    \label{fig:examined-DNN-achitectures}
\end{figure*}

\subsection{Building blocks for feature extraction}
We give special emphasis on the building blocks used to design the DNN model that performs the device mapping classification task (\circled{4}).
Compared to Deeptune, we incorporate three additional processing layers, that aim to tackle Deeptune's limitations presented above.

First,  we extend DeepTune by adding a CNN layer, that receives as input the output of the embedding layer.
This method takes advantage of the inherent structure of a textual label by identifying common attributes shared by multiple words and dividing them into three equal parts called trigrams~\cite{poznanski2016cnn}.
Trigrams, we believe, are an effective representation of programming syntax structure and helps mitigating the gap of remark \textbf{(R1)}.
It is composed of 2 layers: one 1-d convolutional layer, with 64 filters, kernel size of 3 and striding step 1, and one max pooling layer of size 4. 
The embedding layer transmits the words to the convolutional layers in the form of sentences (in our case code instructions). Convolution layer convolve the input using pooling layers; pooling layers aid in reducing the representation of input phrases, input parameters, computation, and overfitting in the network. 

 For preserving both past and future information on the input data (remark \textbf{R2}), we employ bidirectional LSTM layers.
 Specifically, Bi-LSTMs are able to obtain knowledge about the sequence in both directions, backwards (future to past) and forwards (past to future) and recognize long-term dependencies~\cite{schuster1997bidirectional}. 
 The core idea, is to split the state neurons of a regular LSTM in a part that is responsible for the positive time direction (forward states) and a part for the negative time direction (backward states). 
 Outputs from forward states are not connected to inputs of backward states, and vice versa. 
 Then the hidden states of the two LSTMs are combined to find the hidden state for each time point. 
 It is proven to be more accurate than the traditional LSTM networks~\cite{siami2019performance}, but in trade off significant training time. In our case, bidirectional keeps information from both directions, making it easier for the network to understand long-term code dependencies, such as function declarations. 
 
 Last, we tackle the problem of significance-aware feature processing (remark \textbf{R3}) by exploiting a novel solution in the field of NLP. namely Attention layers~\cite{vaswani2017attention}. 
 The attention block highlights the importance of different features that are highly correlated with classification, by assigning weights to features, extracting the contextual information. Attention can be proven to be useful in the case of programming code modeling, emphasizing more critical points, such as branches, loops etc., against less important, such as variable declarations, increments etc. 

\subsection{Examined DNN architectures}
\label{subsec:examined-dnn-archs}
To understand the impact of each additional layer on the overall accuracy of the device mapping problem, we design different DNN architectures with diverse combinations of the aforementioned building blocks, as shown in Fig.~\ref{fig:examined-DNN-achitectures}.
These architectures extend the baseline model of Deeptune (Fig.~\ref{fig:deeptune}) in the following directions:
\begin{itemize}
    \item \textbf{Deeptune-CNN}: Includes a CNN layer right after the Embedding, thus, introducing the aspect of trigrams (Fig.~\ref{fig:cnn}). This model examines the impact of feature extraction from adjacent tokens in the source code (R1).
    \item \textbf{Deeptune-BiLSTM}: We replace the two unidirectional LSTM layers of Deeptune, with a bidirectional one (Fig.~\ref{fig:bilstm}). Through this model, we explore the effect of preserving past and future information in the code (R2).
    \item \textbf{Deeptune-Attention}: This architecture introduces an Attention layer right after the LSTM, which identifies important features in the input. With this model, we explore whether a significance analysis on the features of the input affect the efficiency of the prediction (R3). 
    \item \textbf{Hybrid Architecture}: Final, the hybrid architecture combines all the aforementioned techniques and examines their aggregated impact on the overall accuracy of the model.
\end{itemize}


%% file: sections/evaluation.tex
\section{Evaluation}
\label{sec:evaluation}
We evaluate our proposed methodology and compare it directly with Deeptune by examining the accuracy of each one of the DNN architectures presented in Section~\ref{subsec:examined-dnn-archs}.

\subsection{Examined Dataset}
\par We base our evaluation on the dataset of~\cite{cummins2017end} in order to have accurate comparison between the different model architectures. The dataset contains preprocessed and tokenized OpenCL kernels, from 7 different benchmark suites. Additionally, it contains the execution times for each kernel on two GPU devices, the AMD Tahiti 7970 and the NVIDIA GTX 970, as well as on an Intel Core i7-3820 CPU. Table \ref{tab:platform_cummings} contains details about the CPU-GPU platforms.

The prediction target is the platform in which the execution time is lower. More precisely, when we examine the AMD GPU and Intel CPU cases, the target is $[1,0]$ if the kernel runs faster on the GPU and $[0,1]$ if the kernel runs faster on the CPU. This is also referred to as one-hot encoding. Likewise, for the NVIDIA GPU and the Intel CPU case.

\begin{table}[t]
\centering
\begin{tabular}{llll}
\rowcolor[HTML]{C0C0C0} 
{\color[HTML]{333333} Platform} & Frequency & {\color[HTML]{333333} Memory} & Driver        \\
Intel Core i7-3820              & 3.6 GHz   & 8 GB                          & AMD 1526.3    \\
AMD Tahiti 7970                 & 1000 MHz  & 3 GB                          & AMD 1526.3    \\
NVIDIA GTX 970                  & 1050 MHz  & 4 GB                          & NVIDIA 361.42
\end{tabular}
\caption{Platform Details}
\label{tab:platform_cummings}
\end{table}

\subsection{Experimental Setup}
\label{subsec:experimental-setup}
We use stratified 10-fold cross-validation to evaluate the predictive quality of each model. Each program is randomly assigned to one of ten equal-sized sets; the sets are balanced to ensure a consistent proportion of samples from each class across the whole set. A model is trained on all but one of the sets' programs and then tested on the programs from the unseen set. This procedure is done for each of the ten sets in order to provide a comprehensive prediction for the entire dataset.

All models were implemented in Python using Tensorflow\footnote{Tensorflow: https://www.tensorflow.org/} and Keras\footnote{Keras: https://keras.io/} backends. To ensure the most accurate comparison, we seed our layers with the same number, in order to be initialized with the same weights. The maximum sequence length was set to $1024$ and the learning rate at $10^{-3}$. We used categorical cross entropy as loss function, a batch size of $64$ and trained for $50$ epochs. As optimizer, we used the adaptive learning rate optimization algorithm, Adam. The experiments were carried out on an NVIDIA Tesla V100 GPU. Finally, we used TensorBoard\footnote{TensorBoard: https://www.tensorflow.org/tensorboard/} to measure and visualize parameters like loss and accuracy.

\subsection{Experimental Results}
The average accuracy of each model, as measured in a 10-fold test set, is shown in Table \ref{tab:arch_evaluation}, where the best results are printed in \textbf{bold} font.

\begin{table*}[t]
\centering
\begin{tabular}{|l|l|l|l|}
\hline
{\color[HTML]{333333} } & \textbf{AMD Tahiti 7970} & \textbf{NVIDIA GTX 970} & \textbf{Average}     \\ \hline
\textbf{DeepTune~\cite{cummins2017end}} & $83.23\%$             & $80.29\%$             & $81.76\%$                         \\ \hline
\textbf{DeepTune-CNN}                   & $85.88\%$             & $80.01\%$             & $82.94\%$                         \\ \hline
\textbf{DeepTune-bilstm}                & $84.85\%$             & $80.88\%$             & $82.87\%$                         \\ \hline
\textbf{DeepTune-Attention}             & $83.91\%$             & $81.03\%$             & $82.50\%$                         \\ \hline
\textbf{CNN-BiLSTM-Attention}           & $\mathbf{87.35\%}$    & $\mathbf{81.47\%}$    & $\mathbf{84.41\%}$                \\ \hline
\end{tabular}
\caption{Accuracy per examined DNN architecture (as described in Sec.~\ref{subsec:examined-dnn-archs})}
\label{tab:arch_evaluation}
\end{table*}

\begin{figure}
    \includegraphics[scale=0.58]{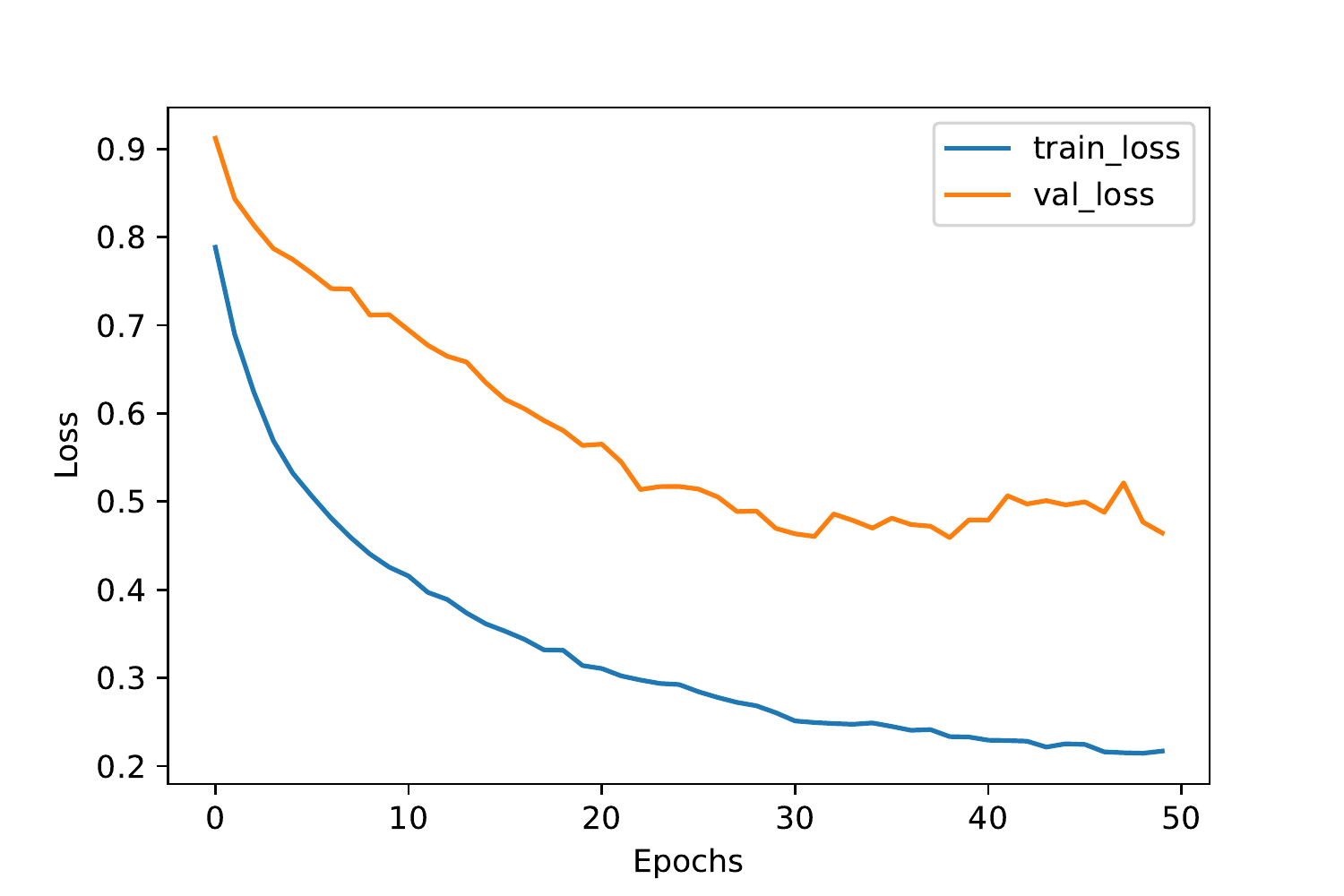}
    \caption{Train and validation loss of the Hybrid Architecture model.}
    \label{fig:proposed_loss}
\end{figure}
\begin{figure}
    \includegraphics[scale=0.58]{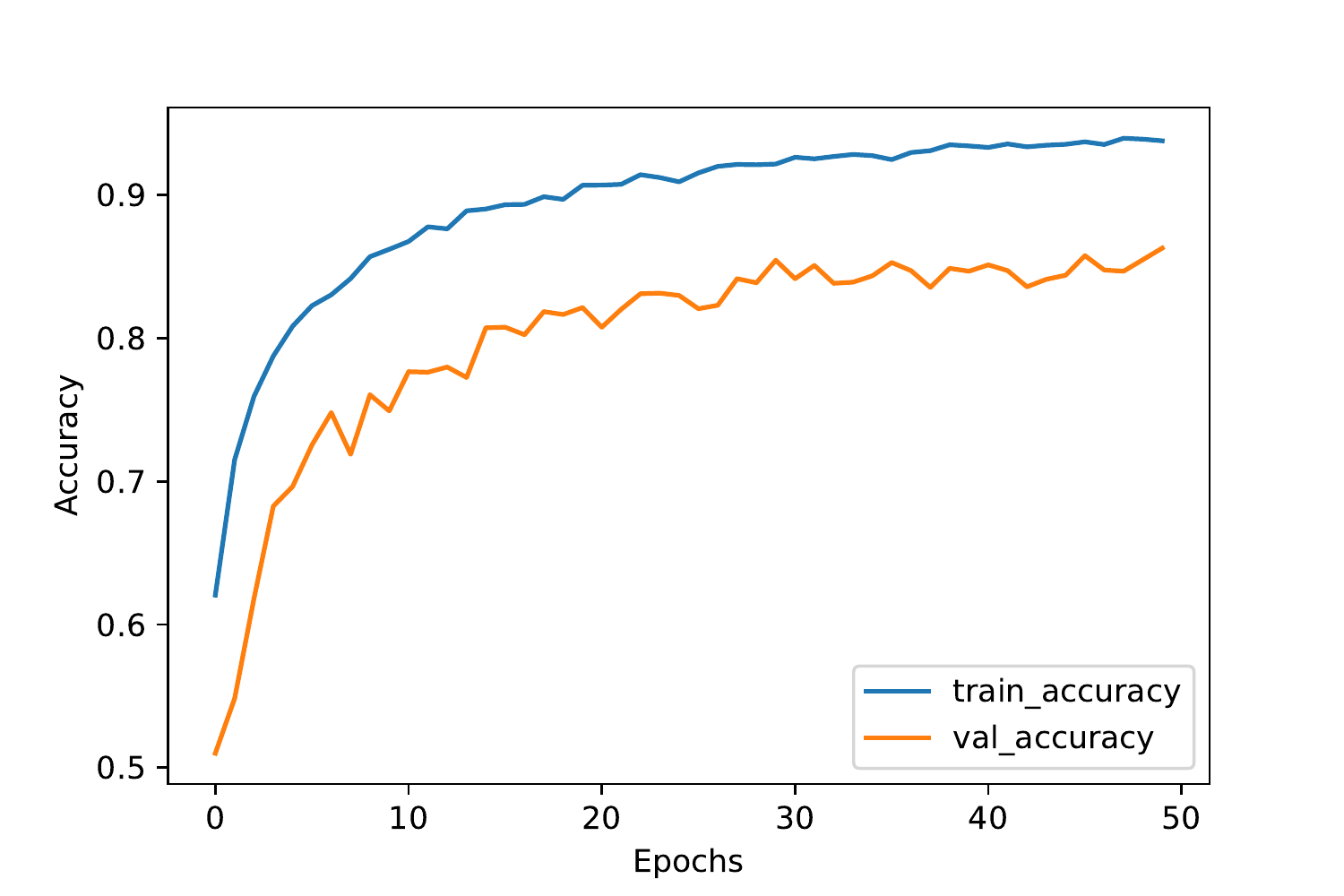}
    \caption{Train and validation accuracy of the Hybrid Architecture model.}
    \label{fig:proposed_acc}
\end{figure}

\textbf{\underline{Baseline architecture:}} The baseline architecture is Deeptune, the model suggested in~\cite{cummins2017end}.
For the shake of fair comparison, we have retrained the model using the steps of Sec. \ref{subsec:experimental-setup}, rather than hard copying the results from \cite{cummins2017end}.
We observe that Deeptune achieves an average accuracy of $81.76\%$, with better results on the AMD platform with $83.23\%$, while in the NVIDIA device it scored $80.29\%$. 

\textbf{\underline{Impact of Trigrams (R1):}} We observe that the notion of trigrams helps the model's code comprehension capabilities.
Specifically, the Deeptune-CNN model already outperforms the baseline by $2.65\%$ on AMD platform and $1.18\%$ on average, but performs slightly worse on the NVIDIA platform ($0.28\%$ accuracy drop), which, however, is statistically insignificant.
We also observe that the introduction of the CNN layer provides the greatest accuracy increment in the case of AMD GPU compared to that of Bi-LSTM and Attention.

\textbf{\underline{Impact of preserving past and future information (R2):}} 
The Deeptune-BiLSTM model, performs better in both devices compared to Deeptune. 
More specifically the model scored $84.85\%$ on the AMD device and $80.88\%$ on the NVIDIA, resulting to an average precision of $82.87\%$, that is $1.11\%$ improvement than the baseline.
Moreover, we notice that similar to the case of trigrams, the accuracy increment is lower for the NVIDIA case.

\textbf{\underline{Impact of feature significance (R3):}} The addition of an attention layer in DeepTune-Attention model, also outperforms the baseline in both devices, scoring $83.91\%$ on the AMD platform and $81.03\%$ on the NVIDIA, leading to $82.50\%$ on average. It appears to give the best results, so far, on the NVIDIA platform, that is $0.73\%$ higher than the baseline.

\textbf{\underline{Hybrid Architecture:}} Last, our proposed hybrid model, CNN-BiLSTM-Attention, outperforms the baseline model and all other models.
This reveals that by combining all of our proposed feature processing steps, we can obtain much greater contextual information from the processed kernel and, thus, maximize the accuracy of our model.
Specifically, the hybrid architecture scored $87.35\%$ on the AMD platform and $81.47\%$ on the NVIDIA device, that is $4.12\%$ and $1.18\%$ higher than the baseline, respectively, resulting in an average accuracy increase of $2.65\%$, that is $84.41\%$.

For the shake of completeness, we also report the loss and accuracy curves over the training phase, for the Hybrid Architecture.
Figures \ref{fig:proposed_loss} and \ref{fig:proposed_acc} show the respective results, where each curve corresponds to the average loss and accuracy over the 10-fold cross train and validation sets. 
As can be seen, the validation loss decreases until epoch 40, at which point it remains pretty constant. Because the training loss continues to decrease, we stop training until epoch 50 to avoid overfitting~\cite{hawkins2004problem}.

%% file: sections/conclusion.tex
\section{Conclusion}
\label{sec:conclusion}
In the era of extreme device heterogeneity, choosing the most appropriate device for application execution forms a really challenging problem.
In this paper, we build upon the work of Cummins et al. \cite{cummins2017end}, and pinpoint some of its limitations.
Then, we examine the impact of additional feature extraction approaches for improving the accuracy of optimal device mapping prediction for OpenCL kernels.
On average, our suggested model, namely CNN-BiLSTM-Attention, outperforms the baseline model, surpassing it on both prediction challenges. We achieve \(4.12\%\) higher prediction accuracy than the baseline on the AMD platform and \(1.18\%\) higher on the NVIDIA platform.

In future work, we will expand our method to source code modeling by using transformers to do unsupervised training on unlabeled C code in order to enhance programming language understanding~\cite{devlin2018bert}; continue our investigation into the optimization of CUDA kernels; and train models to aid in the development of energy-efficient embedded devices.


\section*{Acknowledgment}
This work is partially funded by the EU Horizon 2020 research and
innovation programme, under project EVOLVE, grant agreement No 825061.